\documentclass{article}

\usepackage[preprint]{neurips_2026}

\usepackage[utf8]{inputenc}
\usepackage[T1]{fontenc}
\usepackage{hyperref}
\usepackage{url}
\usepackage{booktabs}
\usepackage{amsfonts}
\usepackage{amsmath}
\usepackage{nicefrac}
\usepackage{microtype}
\usepackage{xcolor}
\usepackage{graphicx}
\usepackage{multirow}

\title{EditPropBench: Measuring Factual Edit Propagation in Scientific Manuscripts}
\author{%
  Garvin Kruthof \\
  Technical University of Munich \\
  \texttt{garvin.kruthof@tum.de}
}

\begin{document}

\maketitle
\begin{abstract}
Local factual edits in scientific manuscripts often create non-local revision obligations. If a dataset changes from 215 to 80 documents, claims such as ``medium-scale'' or ``a few hundred items'' may also become stale, even though they do not repeat the edited number. In an audit of recent arXiv cs.CL benchmark and dataset papers, we find fact-dependent qualitative claims in 37.2\% of papers, suggesting that this dependency pattern is common in the target genre. We introduce EditPropBench, a benchmark for measuring whether LLM editors propagate factual edits through dependent manuscript claims. Each item contains an ML/NLP-style synthetic manuscript, a targeted edit, and a controlled fact graph with sentence-level labels for direct targets, required downstream updates, and unrelated text that should remain unchanged. We summarize cascade success with Edit-Ripple Adherence (ERA), the fraction of required downstream updates correctly revised, and validate the metric with adversarial probes and stress-test variants. On the hardest cases, where dependent claims use implicit or free-form wording rather than repeating the edited value, five LLM editing systems span ERA 0.148-0.705. Even the strongest misses roughly 30\% of required cascade updates. This advantage persists in a mixed evaluation that includes easy cases solvable by deterministic substitution. EditPropBench shows that current LLM editors can repair many implicit consequences of factual edits, but reliable scientific revision still requires cascade-aware checking.
\end{abstract}
\section{Introduction}
\label{sec:intro}

LLM writing tools are increasingly used to revise research manuscripts. A useful editor must do more than replace a local string: when a factual value changes, every claim licensed by that value may need to change as well. For example, changing a dataset size from ``215 documents'' to ``80 documents'' may require revising qualitative descriptions such as ``medium-scale,'' ``neither tiny nor sprawling,'' or ``a few hundred items.''

We study this revision problem as \emph{cascade propagation}: propagating a local factual edit through all manuscript sentences whose meaning depends on the edited fact. This failure is easy to miss. A model may correctly rewrite the explicit number while leaving dependent qualitative claims untouched. Surface-level diff checks would mark the local edit as successful, but the revised manuscript would remain internally inconsistent.

Existing benchmarks do not directly measure this capability. Knowledge-editing benchmarks evaluate ripple effects over triples, typically through short factoid questions. Document-level editing benchmarks study larger revisions, but usually lack a controlled fact graph specifying which sentences should change and why. Factual-consistency benchmarks detect contradictions after generation, but do not test whether an editor can repair all dependent claims after a targeted edit. EditPropBench fills this gap by combining controlled fact graphs with long-form scientific prose and sentence-level dependency labels.

\paragraph{What EditPropBench measures.}
Each item pairs a manuscript with a local edit and sentence-level labels identifying direct targets, required downstream updates, and protected unrelated units. The headline metric, Edit-Ripple Adherence (ERA), measures whether required updates were successfully revised. We stratify dependent sentences by whether the dependency is explicit, literal, implicit, or free-form; the latter two form the hard stratum, where simple substitution rules cannot solve the task.

\paragraph{Contributions.}
We make four contributions. First, we introduce EditPropBench, a controlled manuscript-level benchmark with 1{,}907 fact-dependent sentence-level targets and 461 hard implicit/free-form scoring cells. Each item includes a fact graph, sentence-level dependency labels, three editing protocols, and ERA-based scoring. Second, we show that LLM editors substantially outperform substitution baselines on implicit and free-form dependencies, but remain unreliable: the strongest system reaches ERA 0.705 on the hard stratum. Third, we validate the metric with adversarial probes, judge swaps, a human pilot, and a stress-test corpus that includes substitution-solvable cases. Fourth, we audit recent arXiv cs.CL benchmark and dataset papers and find that fact-dependent qualitative claims are common in scientific writing.

\section{Related Work}
\label{sec:related}

\paragraph{Scientific and document-level revision.}
Automated academic writing assistance has studied revision from rough-draft rewriting and interactive editing systems to scientific-article revision corpora~\citep{ito-etal-2019-diamonds, ito-etal-2020-langsmith, jourdan-etal-2024-casimir, jourdan-etal-2025-pararev}. General instruction-based editing and document-level model-editing benchmarks further move beyond isolated sentences toward longer revision contexts~\citep{Raheja2023CoEdITTE, Zeng2025DocMEditTD, Rosati2024LongformEO}. These resources better match manuscript editing than short factual probes, but they generally do not provide an exhaustive fact-to-sentence dependency graph. EditPropBench trades some surface realism for controlled supervision: each sentence is linked to the facts it depends on, allowing us to measure whether a local factual edit is propagated to all dependent manuscript claims, including implicit qualitative descriptions.

\paragraph{Knowledge editing and ripple effects.}
Knowledge-editing methods and benchmarks study how to change a model's behavior on a target fact while preserving unrelated knowledge~\citep{de-cao-etal-2021-editing, Meng2022LocatingAE, Meng2023MEMIT, Mitchell2022MemoryBasedME, Yao2023EditingLL, Wang2024EasyEdit, Zhang2024ACS, Akyrek2023DUnEDF, Chen2025UniEditAU}. Ripple-effect evaluations are closest in spirit because they ask whether edits affect related facts rather than only the directly edited one~\citep{Cohen2023EvaluatingTR, Zhong2023MQuAKEAK}. EditPropBench adopts this consequence-oriented view but changes the output modality: instead of probing edited model knowledge with QA or cloze-style prompts, it evaluates whether an editor revises a full scientific manuscript so that all fact-dependent textual claims remain internally consistent. Appendix~\ref{sec:appendix-related-comparison} positions EditPropBench against the closest prior work on five axes: output unit, dependency labels, qualitative-descriptor coverage, manuscript-level scope, and judge-assisted scoring.

\paragraph{Evaluation, contradiction detection, and drift.}
Scientific-revision and contradiction-detection work shows that evaluation depends strongly on corpus construction, alignment, reference coverage, and metric choice~\citep{Jourdan2025IdentifyingRE, Chen2025ThinkWD, Kryscinski2019EvaluatingTF, Laban2021SummaCRN, Hou2024WikiContradictAB}. We therefore report both a hard-stratum analysis isolating implicit and free-form dependencies and a stress-test aggregate that includes easier cases. Because valid cascade updates may be natural paraphrases, we use reference-guided LLM judging after deterministic matching, following prior LLM-evaluation methodology~\citep{Zheng2023JudgingLW, Bai2024MTBench101AF, Kwan2024MTEvalAM, Sirdeshmukh2025MultiChallengeAR}.

\section{Benchmark Design}
\label{sec:design}

\subsection{Task formulation}

Each item consists of a manuscript $M$, a local edit instruction $e$, and sentence-level annotations. The edit instruction specifies a target fact, its old value, and its new value. The annotations partition manuscript sentences, which we call \emph{units}, into three sets: \textit{direct-target units}, sentences that explicitly mention the edited fact and must use the new value; \textit{required-update units}, sentences whose meaning depends on the edited fact and must also be revised; and \textit{protected units}, sentences unrelated to the edited fact that should remain unchanged.

Given a candidate revision $M'$, the scorer computes three primary metrics. \textit{Local Edit Success} (LES) is the fraction of direct-target units correctly updated. \textit{Edit-Ripple Adherence} (ERA) is the fraction of required-update units correctly updated and is the headline cascade-propagation metric. \textit{Collateral Damage Rate} (CDR) is the fraction of protected units damaged by the revision. We also report a global consistency score (GCS) and a coherence-conditioned gap,
\[
\text{CCG}=\max(0,\text{LES}-\text{GCS})\cdot \sigma((\text{LES}-0.7)/0.05),
\]
where $\sigma$ is the logistic sigmoid. The sigmoid term is a soft local-edit-success gate: it is near zero for systems that rarely perform the direct edit, equals $0.5$ at LES $=0.7$, and approaches one once LES is comfortably above that threshold. \footnote{We use a width of $0.05$ so the transition occurs over a narrow but continuous band rather than as a hard cutoff. This prevents systems that simply avoid editing from appearing globally coherent.}

\subsection{Dependent sentence types}

Required-update units are stratified into four types, ordered from easiest to hardest for deterministic baselines:
\begin{itemize}
  \item \textit{Number-mention}: the sentence states the edited value directly, e.g., ``the corpus contains 230 documents.''
  \item \textit{Literal-qualifier}: the sentence uses the canonical qualitative label associated with the value, e.g., ``a large-scale benchmark.''
  \item \textit{Implicit}: the sentence uses a paraphrased qualifier from a small closed set, e.g., ``of intermediate scope.''
  \item \textit{Free-form}: the sentence uses an open-ended paraphrase, e.g., ``occupying a workable middle ground in size.''
\end{itemize}
The union of implicit and free-form units is the \textbf{hard stratum}. This stratum most directly tests semantic cascade propagation rather than string replacement. Aggregating all unit types hides the failure mode: systems are near ceiling on number mentions and literal qualifiers, while still missing many implicit and free-form dependencies.

\subsection{Worked example}

\begin{figure}[h]
\centering
\includegraphics[width=0.92\linewidth]{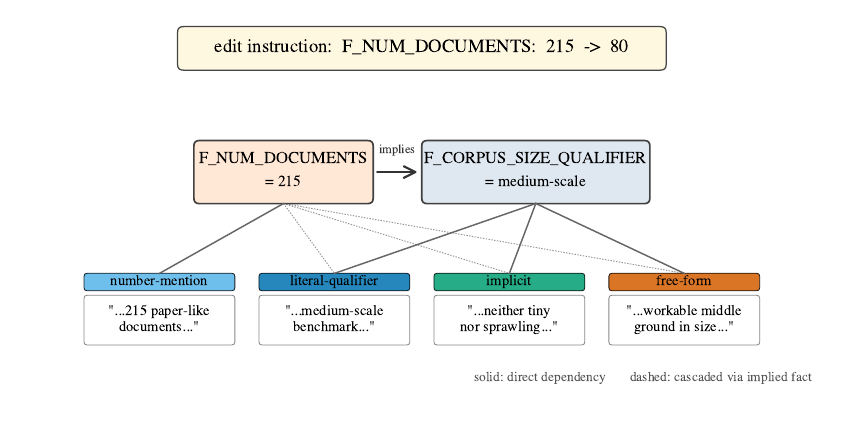}
\caption{Fact graph for one EditPropBench item. An edit on the numeric fact (top) implies an update to the derived qualifier fact, which in turn requires updates to four sentences -one per dependent-sentence type. Solid lines mark direct fact-to-sentence dependencies; dashed lines mark cascaded dependencies that flow through the implied qualifier fact.}
\label{fig:fact_graph}
\end{figure}

In one benchmark item, the edit changes a manuscript's document count from $215$ to $80$ and changes the coupled corpus-size qualifier from \emph{medium-scale} to \emph{small-scale}. The direct target names the count. Several other required-update units depend on the same fact:
\begin{itemize}
  \item \textit{Number-mention}: ``This paper presents ManuscriptRippleBench, evaluating edit propagation across 215 paper-like documents.''
  \item \textit{Literal-qualifier}: ``We position ManuscriptRippleBench as a medium-scale evaluation suite for revision-consistency phenomena.''
  \item \textit{Free-form}: ``We treat ManuscriptRippleBench as a benchmark neither tiny nor sprawling, complementing larger natural-document evaluations.''
  \item \textit{Free-form}: ``Our experiments used a setup spanning a few hundred items, so extending to larger systems will require additional benchmarks.''
\end{itemize}
A model that edits only the explicit count succeeds locally but fails globally. After the edit, ``neither tiny nor sprawling'' and ``spanning a few hundred items'' no longer fit an 80-document corpus. The task is therefore not reducible to value replacement: the revision requires recognizing that a qualitative description is licensed by a quantitative fact. Figure~\ref{fig:fact_graph} shows the corresponding fact graph: the numeric fact \texttt{F\_NUM\_DOCUMENTS} implies the qualifier fact \texttt{F\_CORPUS\_SIZE\_QUALIFIER}, and an edit on the numeric fact must therefore propagate through both directly attached and cascade-attached sentences.

\subsection{Corpus construction and editing protocols}

\begin{figure}[h]
\centering
\includegraphics[width=0.85\linewidth]{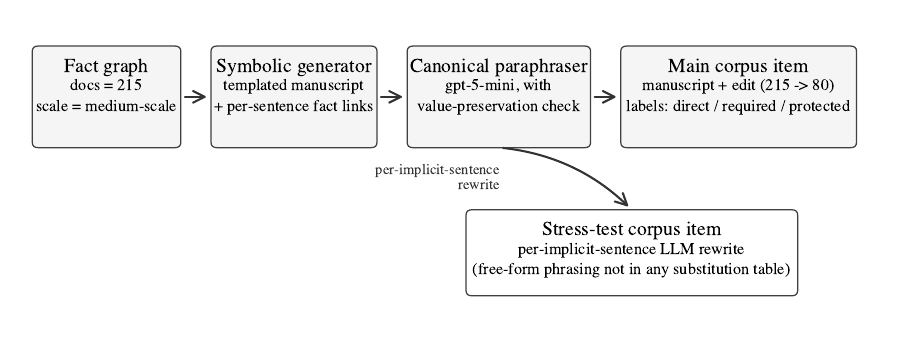}
\caption{Corpus construction pipeline. The symbolic generator emits a templated manuscript from a fact graph; the canonical paraphraser rewrites each sentence with a value-preservation check; pairing with an edit instruction yields a main-corpus item with per-sentence type labels. A second per-implicit-sentence LLM rewrite produces the stress-test corpus item used for cross-validation in Section~\ref{sec:stress}.}
\label{fig:pipeline}
\end{figure}

EditPropBench manuscripts are generated in three stages (Figure~\ref{fig:pipeline}). First, a symbolic generator emits ML/NLP-style manuscripts from a fact graph and records which sentences depend on which facts. Second, a fixed canonical paraphraser, \texttt{gpt-5-mini} snapshot 2026-01-15, rewrites each generated sentence. Validation checks ensure that required fact values are preserved. This pass turns rigid template text into more natural prose and introduces implicit and free-form dependencies. Third, the paraphrased manuscripts are paired with edit instructions that change one fact and induce a known set of direct targets, required updates, and protected units.

We release two corpus variants. The main corpus contains 122 paraphrased manuscripts and supports the headline hard-stratum analysis. The stress-test corpus rewrites each implicit sentence through a separate LLM call, producing unique free-form phrasing that is not enumerable by a substitution table. This variant tests whether the cascade gap persists when deterministic baselines are not pinned to zero on the aggregate. We evaluate three protocols: \textit{direct edit}, where the model returns the full revised manuscript; \textit{patch edit}, where it returns only rewritten sentences with unit ids; and \textit{plan-then-edit}, where it emits a short plan before the full revision.

\paragraph{Construct realism.}
The synthetic construction is deliberate. Fully natural manuscripts rarely expose the latent dependency graph needed to score recall over all affected claims. EditPropBench instead targets \emph{construct realism}: it contains the same kind of fact-to-claim dependency that occurs in scientific writing, while retaining exhaustive ground truth for which sentences should and should not change after a counterfactual edit. Section~\ref{sec:natural_evidence} tests whether this dependency pattern appears in recent scientific prose.

\paragraph{Corpus statistics.}
The 122 benchmark items span 89 distinct manuscripts (1.37 items per manuscript on average; some manuscripts appear with two edits). Each item carries a mean of $15.6$ required-update units (median $16$, range $10$--$21$), totalling $1{,}907$ dependent sentences across the corpus. Required-update units stratify as $1{,}239$ number-mention ($\sim 65\%$), $226$ literal-qualifier ($\sim 12\%$), $168$ implicit ($\sim 9\%$), and $274$ free-form ($\sim 14\%$). The union of implicit and free-form forms the $442$-unit hard required stratum; the scorer additionally evaluates $19$ deletion-tagged units in the same hard semantic category, giving a hard-stratum scoring denominator of $n{=}461$ for the headline ERA. Per item, hard cells (req + deletion) have mean $3.78$ (median $4$, range $1$--$7$); no item has zero hard cells, so every benchmark item contributes to the headline metric. Manuscript-clustered bootstrap CIs reported in Section~\ref{sec:results} resample the $89$ manuscripts to account for within-manuscript dependence.

\section{Scoring and Validation}
\label{sec:scoring}

\subsection{Deterministic and judge-assisted scoring}

Scoring has two stages. \textit{Stage A} is a deterministic matcher. For required-update units, it credits a revision when the new value or a registered alias is present and the old value is absent, with exceptions for valid revision-marker references to the old value. This handles explicit number mentions and many literal-qualifier cases. \textit{Stage B} is a three-call LLM-judge ensemble (\texttt{claude-sonnet-4-6}) used as a fallback when Stage A returns INCORRECT or MISSING. The judge is asked whether the revised unit semantically asserts the new fact value, including via natural-paraphrase substitutions outside the canonical alias list (e.g., ``with narrow coverage of the model space'' for new value \textsc{narrow-pool}). Stage B is also used to grade protected-unit damage.

Both required and protected scoring use the same Sonnet ensemble. We report the combined Stage A + B verdict as the headline metric. This choice is load-bearing on the hard stratum because Stage A is deliberately conservative: it cannot credit a semantically correct rewrite that does not surface a registered alias. Diagnostic inspection showed that roughly two-thirds ($295$ of the $431$ hard cells where Stage A did not return SATISFIED, $=68.4\%$) were natural-paraphrase substitutions a careful reader would credit as successful propagation. Appendix~\ref{sec:appendix-examples} catalogues the per-unit decision categories with concrete examples.

\subsection{Falsification gate}

Before interpreting model results, we run deterministic adversarial baselines designed to reveal metric exploits. Identity returns the input unchanged; Half-Identity and Three-Quarter-Identity rewrite only some required units; Null returns an empty document. Hedge-Bot variants insert hedge clauses without making genuine changes, and Truncation-Abuse returns only an early portion of the manuscript. The Edit-Direct-Target-Only probe edits exactly the units that explicitly mention the edited fact value using word-boundary substitution and copies every other unit verbatim. These probes test whether the metric can be gamed by doing nothing, editing only partially, hedging, omitting difficult material, or applying a literal substitution rule without cascade reasoning. Each probe has a preregistered paired-bootstrap CI bound. Verdicts are \texttt{pass}, \texttt{fail}, or \texttt{indeterminate}; the latter is used when LES is too low for the CCG gate to make the comparison informative.

\section{Experiments and Results}
\label{sec:results}

\subsection{LLM editors improve hard-stratum cascade propagation but do not solve it}

\begin{figure}[h]
\centering
\includegraphics[width=0.90\linewidth]{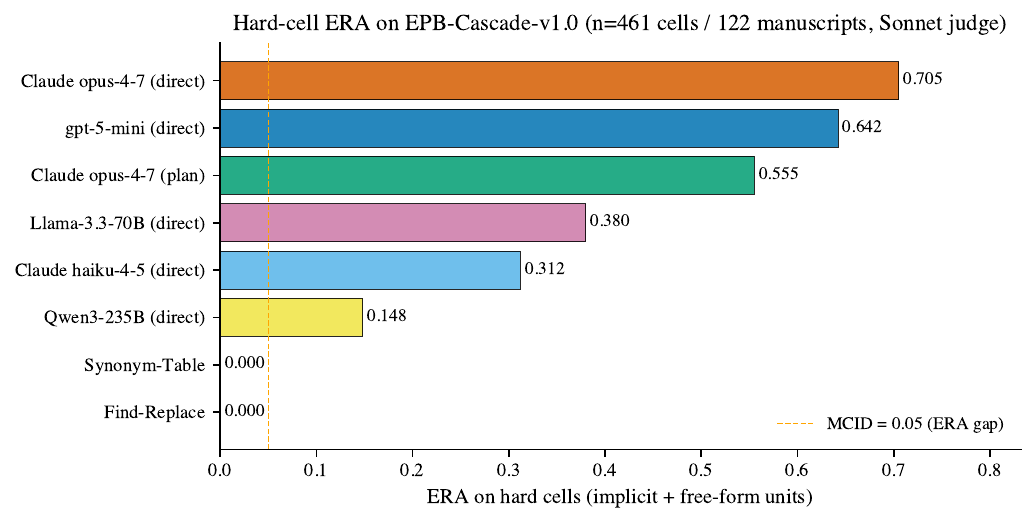}
\caption{Hard-stratum ERA on the main corpus ($n=461$ implicit + free-form dependent sentences across 122 manuscripts), under the Sonnet-judged Stage A + B scorer. Find-Replace and Synonym-Table score $0\%$ by construction. The dashed reference line at $0.05$ corresponds to the preregistered minimum important difference for ERA gaps.}
\label{fig:hero}
\end{figure}

On the main corpus hard stratum ($n=461$ implicit and free-form cells across 89 manuscripts), the five evaluated direct-edit systems span ERA $0.148$ to $0.705$ (Figure~\ref{fig:hero}). The best model is $\sim 4.8\times$ stronger than the weakest. Plan-then-edit underperforms direct-edit for opus ($0.555$ vs.\ $0.705$). For construct isolation, opus direct lifts ERA by $+70.5$ pp over Find-Replace and Synonym-Table (95\% manuscript-clustered paired bootstrap CI $[+63.3,+77.5]$, $B=2000$); this is a floor sanity check because the stratum excludes units those baselines can solve. The more comparable mixed-stratum effect-size estimate is the stress-test aggregate gap in Section~\ref{sec:stress}.

\begin{figure}[h]
\centering
\includegraphics[width=0.92\linewidth]{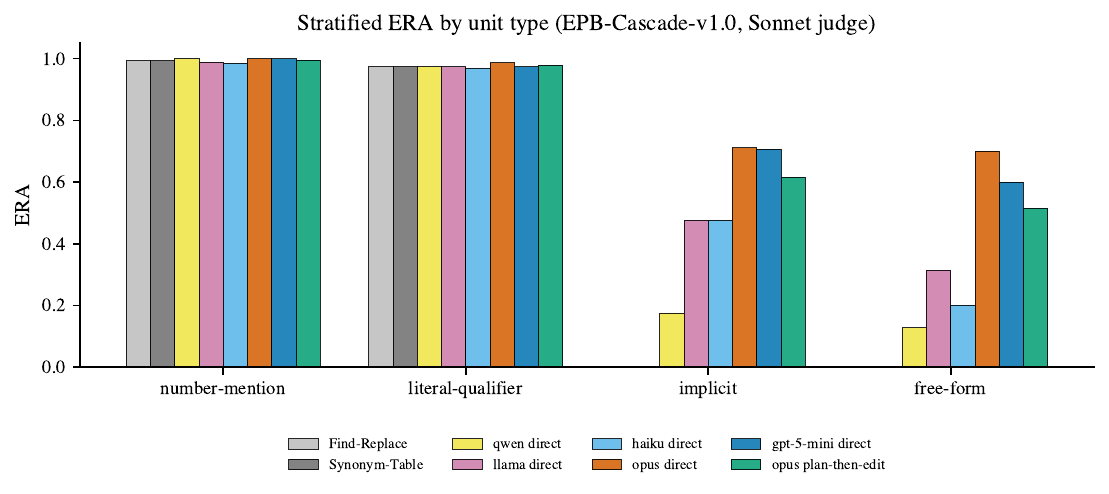}
\caption{ERA by dependent-sentence type. Models are near ceiling on number-mention and literal-qualifier units. The meaningful spread appears on implicit and free-form units, where the dependency is not recoverable by simple substitution.}
\label{fig:strata}
\end{figure}

Two findings emerge. First, the task is feasible: LLMs identify and repair many implicit dependencies, including paraphrases that are not in the canonical alias list. Hard-stratum ERA is computed with the LLM-judge fallback described in Section~\ref{sec:scoring}, which credits semantic substitutions such as ``with narrow coverage of the model space'' for the new value \textsc{narrow-pool}. Second, the task is not solved: even the strongest system misses roughly $30\%$ of hard-stratum cascades under judge-assisted scoring, and semantic update success varies widely across models. Figure~\ref{fig:hero} shows the hard-stratum headline. Figure~\ref{fig:strata} shows why this view is necessary: number-mention and literal-qualifier units are near ceiling, while implicit and free-form units create the meaningful spread.

\subsection{The cascade gap survives stress-test cross-validation}
\label{sec:stress}

The hard-stratum analysis isolates the construct of interest, but it also pins deterministic substitution baselines to zero by excluding the units they can solve. We therefore run a stress-test analysis that restores number-mention and literal-qualifier units to the denominator while rewriting implicit dependencies so that they are not enumerable by a substitution table. This mixed corpus asks whether the LLM advantage remains visible when easy substitution-solvable cases are included.

On this aggregate, Find-Replace and Synonym-Table reach ERA $0.7504$ by recovering the easy strata. All four evaluated LLM cells exceed that floor: \texttt{gpt-5-mini} direct reaches $0.904$ ($+15.4$ pp, item-clustered 95\% CI $[+13.6,+17.3]$), \texttt{Llama-3.3-70B} direct reaches $0.867$ ($+11.7$ pp), \texttt{Qwen3-235B} plan-then-edit reaches $0.827$ ($+7.7$ pp), and \texttt{haiku-4-5} patch reaches $0.773$ ($+2.2$ pp, CI $[+1.2,+3.3]$). The smaller gap is expected: easy strata dilute the implicit/free-form cases. The key result is that the advantage remains positive even after substitution-solvable units are included. Appendix~\ref{sec:appendix-stress} plots the hard-stratum and mixed-stratum contrasts side by side.

\subsection{Validation checks support the main interpretation}

Adversarial baselines clear the preregistered falsification gates: identity, partial-edit, hedge, null, and direct-target-only probes do not obtain spurious cascade credit on the cascade corpus; Truncation-Abuse passes on the pilot and scaled corpora and is indeterminate at the cascade scale (LES too low for the CCG gate to be informative; Appendix~\ref{sec:appendix-gate}). The direct-target-only probe reaches aggregate ERA $0.041$ versus Find-Replace's $0.750$, separating local substitution from cascade propagation.

The Stage B judge is robust across calibration checks. Three disjoint judge families clear the held-out probe threshold, and a Llama-based judge swap preserves the ranking of re-scored model cells. A 64-item human pilot supports the load-bearing hard-stratum did-change judgment: Sonnet agrees with the independent external rater at $\kappa=0.871$ on hard implicit/free-form units (one-sided 95\% bootstrap LB $\geq 0.709$). 

\section{Evidence from Real Scientific Manuscripts}
\label{sec:natural_evidence}

EditPropBench uses synthetic manuscripts to obtain complete dependency labels, but the benchmark is useful only if real papers also contain qualitative claims whose wording depends on quantitative facts. To test this assumption, we audit recent arXiv \texttt{cs.CL} benchmark and dataset papers sampled from April 2026 (arXiv IDs \texttt{2604.*}). After PDF parsing and body-text filtering, the audit contains $199$ papers and $4{,}696$ paragraphs.

We use a \texttt{gpt-5-mini} classifier to identify paragraphs where a quantitative fact supports a qualitative claim, such as a dataset size being described as ``large-scale'' or an accuracy value as ``near-perfect.'' This pattern appears in $74$ of $199$ papers ($37.2\%$, Wilson 95\% CI $[30.8,44.1]$), suggesting that fact-dependent qualitative language is common in the target genre.

\section{Discussion and Limitations}
\label{sec:discussion}

\paragraph{Reading the two effect sizes.}
The hard-stratum and stress-test results answer different questions. The hard-stratum gap estimates performance on implicit and free-form dependencies, where simple substitution cannot solve the task. The stress-test aggregate estimates how much that capability remains visible in a mixed corpus dominated by easier cases. Reporting both avoids two errors: over-claiming from a hand-picked hard slice and under-claiming from an aggregate where deterministic baselines already solve many units.

\paragraph{What the benchmark says about current editors.}
The main result is not that LLMs fail to edit manuscripts. They often perform the direct edit, and they can repair some implicit consequences. The result is more specific: current systems do not reliably discover all claims that become stale after a local factual intervention. This distinction matters for deployment. A manuscript can look revised while still containing claims that remain semantically tied to the old fact.

\paragraph{Implications for writing tools.}
EditPropBench is not intended to rank models on general scientific editing. It isolates one revision capability: updating implicit claims that depend on a local factual edit. A writing assistant that passes EditPropBench-style checks would be more trustworthy for manuscript revision, but the present results argue against fully automatic use: even the strongest system misses roughly $30\%$ of hard-stratum cascades under judge-assisted scoring, and weaker mid-tier models miss most. Practical systems should pair generation with explicit cascade detection, structured dependency tracking, or human review.

\paragraph{Limitations.}
First, EditPropBench prioritizes controlled fact graphs over surface indistinguishability from real arXiv prose. We deliberately do not aim to produce manuscripts that pass for real papers under a discriminator because such surface indistinguishability would conflict with exhaustive, mechanically verifiable cascade supervision. Our claim is narrower: the benchmark isolates a dependency pattern that occurs in real manuscripts (Section~\ref{sec:natural_evidence}) and makes that pattern measurable under complete labels. Second, the body-text audit establishes that the targeted dependency pattern is present in roughly one in three current ML/NLP benchmark/dataset papers, but it does not measure full transfer from synthetic manuscripts to real-paper editing; a manually annotated real-paper editing slice with rater-coded propagation outcomes is a natural next validation step. Third, human validation is preliminary: the 64-item pilot supports the Stage B judge as a reasonable substitute on this sample, but does not establish final construct validity. Fourth, we exclude value-laden edit families such as \texttt{claim\_weakening}, where a model might reasonably refuse to silently weaken a claim. Fifth, the generator targets ML/NLP-style benchmark and dataset manuscripts; other domains may differ. Sixth, the v1.0 scorer does not credit $17/461$ ($3.7\%$) deletion-tagged hard cells whose semantically correct realization is absence; treating these as successes would shift the headline from $0.705$ to $0.744$ and does not change the qualitative ranking. We retain the conservative rubric for this submission and slate the credit-for-absence rule for v1.1. Finally, the deterministic baselines are metric probes and reference floors, not strong cascade-aware editing systems. They establish that hard-stratum success is not achievable by enumerable substitution and that the scorer is not gamed by hedging, partial editing, or truncation. Stronger wrapper systems, such as check-then-revise agents or extract-classify-rewrite pipelines, are natural follow-up systems to evaluate with EditPropBench, but are outside this benchmark paper's scope.

\section{Broader Impact and Release}
\label{sec:release}

EditPropBench can improve LLM-assisted writing tools by identifying a concrete failure mode before deployment. The same information could be misused to choose models more likely to produce silent academic mis-revisions, although we do not believe the benchmark reveals capabilities beyond public model behavior. Because the benchmark targets ML/NLP manuscripts, results should not be generalized to safety-critical domains such as medicine or law without additional evaluation.

The benchmark release covers two endpoints: (i)~a HuggingFace dataset
(\url{https://huggingface.co/datasets/EditCoherenceBench/EditCoherenceBench},
$\sim 210$ MB) containing the synthetic main and stress-test corpora,
deterministic baseline and oracle outputs, judge-scored verdicts under the
canonical Sonnet Stage A + B scorer, the natural-manuscript audit outputs
(per-paragraph classifier results, per-section rates, paper IDs and
SHA-256 hashes for reproducibility), and the anonymized human-rater
annotations; and (ii)~a GitHub source repository
(\url{https://github.com/kruthof/EditCoherenceBench}, MIT-licensed)
containing the symbolic generator, the deterministic scorer, the adversarial
baselines, the LLM-judge wrapper, the annotation interface, configs,
datasheet, and the unit-test suite. The full pipeline is reproducible from
a fresh clone via \texttt{bash scripts/reproduce.sh~--hf} (no API keys
required; pulls the dataset from HuggingFace and prints the headline-table
numbers).

We do \emph{not} redistribute the cached arXiv PDFs collected during the
natural-manuscript audit: arXiv papers are author-copyrighted and bulk
re-hosting requires per-paper permission. Audit results expose paper IDs
and SHA-256 hashes so any user can re-fetch the source PDFs under their
own arXiv-crawler agreement. We also withhold raw frontier-model API
outputs pending per-provider license review; the released score files
contain the per-unit verdicts that reviewers need to verify the headline
numbers without the raw provider responses. Synthetic corpora and per-cell
scoring artifacts are CC-BY 4.0; human annotation files are CC-BY-NC 4.0.

\paragraph{Ethics, human raters, and IRB.}
The two-rater pilot (\S\ref{sec:results}) involved no human-subjects data; raters annotated synthetic manuscript revisions only. Rater~1 is the first author. Rater~2 is an independent external annotator who participated voluntarily and without compensation. IRB review was not required at our institution because the annotation task uses no human subjects' data. No personally-identifying rater data is released; the on-disk \texttt{rater\_id} field is normalized to ``rater\_1'' / ``rater\_2'' in the public annotation files.

\paragraph{Compute.}
All model inference and judge-assisted scoring used commercial LLM APIs (Anthropic, OpenAI, Together) on a single workstation with no GPU compute. Per-call token counts and USD cost are recorded in the \texttt{inference\_metadata} field of every model-output JSONL we generated; the public release includes this metadata for the baseline and oracle outputs (frontier leaderboard outputs are withheld per provider ToS, so their per-call metadata is not in the released artifacts). The total inference budget for the v1.0 release was approximately 600 USD across roughly 5{,}000 API calls.

\section{Conclusion}
\label{sec:conclusion}

EditPropBench evaluates whether LLM editors propagate local factual edits through dependent qualitative claims in scientific manuscripts. The benchmark combines controlled fact graphs, sentence-level dependency labels, three editing protocols, judge-assisted scoring, adversarial metric probes, stress-test cross-validation, and evidence from recent arXiv cs.CL papers that the targeted dependency pattern occurs in real scientific writing.

The results show both progress and a remaining reliability gap. Five direct-edit systems span ERA $0.148$--$0.705$ on the hard implicit/free-form stratum, and LLM systems retain a positive aggregate advantage over deterministic substitution baselines on the stress-test corpus. However, even the strongest system misses roughly 30\% of hard cascade dependencies. Reliable LLM-assisted scientific revision will therefore require explicit cascade-aware checking, not only stronger local rewriting.

\newpage

\paragraph{Declaration of LLM usage.}
Large language models (Claude Opus 4.6 and GPT-5.4) were used as research tools in three capacities during this work, namely (1)~as subject models and automated judges within the benchmark itself, which is the paper's object of study, (2)~for code assistance during pipeline development and analysis scripting, and (3)~for drafting and iterative revision of the manuscript text. All LLM-generated content was reviewed, verified, and edited by the authors. All numerical claims were independently verified against the raw data. The experimental design, research questions, interpretation of results, and scientific conclusions are the authors' own.

\bibliographystyle{plainnat}
\bibliography{references}

\appendix

\section{Stress-test cross-validation figure}
\label{sec:appendix-stress}

\begin{figure}[h]
\centering
\includegraphics[width=0.74\linewidth]{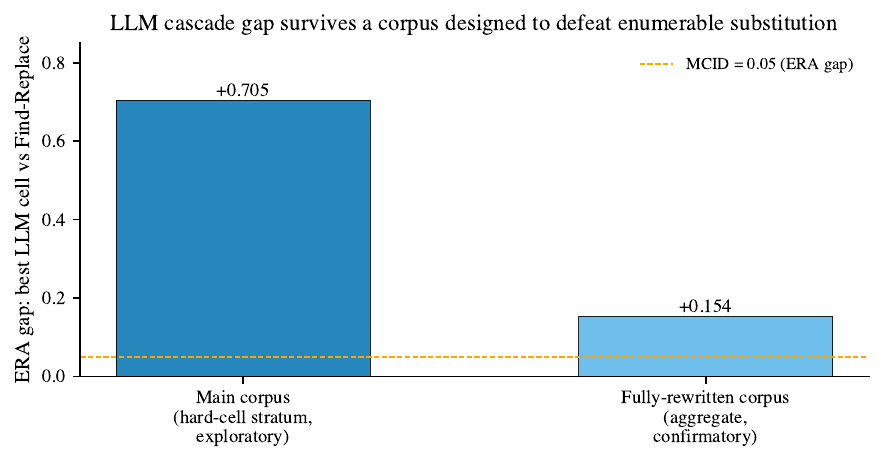}
\caption{Cascade gap for the hard-stratum main-corpus analysis and the stress-test aggregate. The aggregate effect is smaller because easy strata dilute the implicit/free-form cases, but the LLM advantage remains positive on a corpus designed to defeat enumerable substitution rules.}
\label{fig:audit_d}
\end{figure}

Figure~\ref{fig:audit_d} plots the cascade gap on the hard-stratum
main-corpus analysis against the aggregate gap on the stress-test corpus,
in support of Section~\ref{sec:stress}.

\section{Comparison with related benchmarks}
\label{sec:appendix-related-comparison}

Table~\ref{tab:related_comparison} positions EditPropBench against closely related work in ripple-effect evaluation, long-form model editing, and scientific text revision.

\begin{table}[t]
\centering
\scriptsize
\setlength{\tabcolsep}{2pt}
\renewcommand{\arraystretch}{1.08}
\begin{tabular}{p{0.3\linewidth}p{0.17\linewidth}p{0.32\linewidth}cc}
\toprule
Work & Unit & Dependency supervision & Qual. & Doc. \\
\midrule
RippleEdits~\citep{Cohen2023EvaluatingTR} & QA / relation probes & Relation-level ripple labels & -- & -- \\
MQUAKE~\citep{Zhong2023MQuAKEAK} & Multi-hop QA & Multi-hop reasoning chains & -- & -- \\
CASIMIR~\citep{jourdan-etal-2024-casimir} & Article revisions & Aligned edits and revision intentions & -- & \checkmark \\
ParaRev~\citep{jourdan-etal-2025-pararev} & Paragraph rewrite & Revision instructions & -- & -- \\
DocMEdit~\citep{Zeng2025DocMEditTD} & Document edit & Document-level edit facts & -- & \checkmark \\
LEME~\citep{Rosati2024LongformEO} & Long-form generation & Edit-consistency evaluation & -- & partial \\
\textbf{EditPropBench} & \textbf{Manuscript revision} & \textbf{Per-sentence fact graph} & \textbf{\checkmark} & \textbf{\checkmark} \\
\bottomrule
\end{tabular}
\caption{Positioning EditPropBench relative to closely related work. ``Qual.'' indicates whether the benchmark scores implicit qualitative claims whose meaning depends on an edited fact. ``Doc.'' indicates full-document or manuscript-level evaluation.}
\label{tab:related_comparison}
\end{table}

\section{Judge-swap probe-gate calibration}
\label{sec:appendix-probe-gate}

Table~\ref{tab:probe_gate} reports the binary and fine-label accuracy of three Stage B judges against the held-out adversarial probe set. The 60 protected-rubric probes test whether the judge correctly classifies a candidate revision as preserved/benign-rewording vs minor/major-damage; the 20 required-rubric probes test whether the judge correctly identifies satisfied/benign-paraphrase vs incomplete/stale/unrelated propagations of an edited fact value. Binary accuracy is the metric that drives ERA scoring (the binary collapse map in \texttt{DEFAULT\_BINARY\_MAP\_REQUIRED} folds satisfied $\equiv$ benign\_paraphrase and similarly for the protected rubric); fine-label accuracy is reported for diagnostic transparency only.

\begin{table}[h]
\centering
\small
\begin{tabular}{llrrrr}
\toprule
Judge & Rubric & $n$ & Fine acc.\ & Binary acc.\ & Verdict \\
\midrule
Sonnet (primary) & protected & 60 & 0.700 & 0.950 & PASS \\
Sonnet (primary) & required & 20 & 1.000 & 1.000 & PASS \\
Llama-3.3-70B & protected & 60 & 0.750 & 0.900 & PASS \\
Llama-3.3-70B & required & 20 & 0.850 & 1.000 & PASS \\
Qwen3-235B-A22B & protected & 60 & 0.800 & 0.917 & PASS \\
Qwen3-235B-A22B & required & 20 & 0.750 & 1.000 & PASS \\
\bottomrule
\end{tabular}
\caption{Probe-gate calibration: each judge runs the full Stage B protocol (3-call ensemble at $T{=}0$) against held-out adversarial probes (60 protected-rubric, 20 required-rubric). Binary accuracy is the load-bearing metric for ERA scoring (it collapses fine labels under \texttt{DEFAULT\_BINARY\_MAP\_REQUIRED} and the protected analog); fine-label accuracy is reported for diagnostic transparency. The preregistered pass threshold is binary $\geq 0.80$ on each rubric.}
\label{tab:probe_gate}
\end{table}

\section{Plan-then-edit vs direct-edit: per-cell contingency}
\label{sec:appendix-plan-vs-direct}

Table~\ref{tab:plan_vs_direct} reports the per-(item, unit) contingency between opus direct-edit and opus plan-then-edit on the cascade corpus' required-update units, stratified by unit\_type. The headline ERA gap (direct $0.705$ vs plan $0.555$, $\Delta = -0.150$) is concentrated in the hard implicit + free-form stratum: free-form has a regression rate of $0.385$ (74/192 cells where direct-edit succeeded are broken by plan-then-edit) and implicit $0.262$ (32/122). Easy strata (number-mention, literal-qualifier) have regression rates of $0.006$ and $0.018$ respectively. Net cell change in the hard stratum is $-72$ of the total $-78$ across all required-update cells.

\begin{table}[h]
\centering
\small
\begin{tabular}{lrrrrrrr}
\toprule
Stratum & $n$ & BB & BA & AB & AA & reg rate & net \\
\midrule
all required cells & 1932 & 1673 & 120 & 42 & 97 & 0.067 & -78 \\
deletion & 25 & 12 & 3 & 7 & 3 & 0.200 & +4 \\
free-form (hard) & 274 & 118 & 74 & 23 & 59 & 0.385 & -51 \\
implicit (hard) & 168 & 90 & 32 & 11 & 35 & 0.262 & -21 \\
literal-qualifier & 226 & 221 & 4 & 1 & 0 & 0.018 & -3 \\
number-mention & 1239 & 1232 & 7 & 0 & 0 & 0.006 & -7 \\
\bottomrule
\end{tabular}
\caption{Per-(item, unit) contingency between opus direct-edit and opus plan-then-edit on the v1.0 cascade corpus. BB = both pass, BA = direct passes / plan fails (regression), AB = direct fails / plan rescues, AA = both fail. \textit{reg rate} = BA / (BB + BA) — the rate at which plan-then-edit loses ground direct-edit had. \textit{net} = AB $-$ BA. A negative net means plan-then-edit loses more cells than it rescues.}
\label{tab:plan_vs_direct}
\end{table}

\paragraph{Why does plan-then-edit hurt opus?}
\label{sec:appendix-why-ple-opus}

The 15-pp regression of \texttt{opus} plan-then-edit ($0.555$) versus its direct-edit ($0.705$) is concentrated entirely in the hard cells. Per-(item, unit) contingency between the two protocols (Table~\ref{tab:plan_vs_direct}, Appendix~\ref{sec:appendix-plan-vs-direct}) shows that plan-then-edit regresses on $74/192$ ($38.5\%$) of free-form units and $32/122$ ($26.2\%$) of implicit units where direct-edit had succeeded, but only $4/225$ ($1.8\%$) of literal-qualifier units and $7/1239$ ($0.6\%$) of number-mention units. The same regression-rate stratification holds for \texttt{gpt-5-mini} (free-form $22.6\%$, implicit $16.8\%$, easy strata $0\%$): the pattern is not opus-specific. The shape (vanishing on literal cells, large on cells that demand cascade reasoning) is consistent with three non-exclusive mechanisms: planner-induced executor over-conservatism, planner mis-enumeration of \texttt{units\_to\_update}, and edit-call context-budget pressure from the appended plan; the present analysis does not adjudicate among them. \emph{Practical implication.} For cascade-heavy revision workloads, deploy direct-edit; plan-then-edit should be revisited only with planner-side coverage checks against a cascade detector before the edit call.

\section{Stratified pilot $\kappa$ with bootstrap CIs}
\label{sec:appendix-pilot-kappa}

Table~\ref{tab:pilot_kappa_stratified} reports Cohen's $\kappa$ for every cell of the (rater pair $\times$ stratum $\times$ dimension) grid on the 64-item v1.0 human pilot, with two-sided 95\% bootstrap CIs and one-sided 95\% lower bounds. Bootstraps use $B=10000$ item-level resamples, seed $4242$. The hard stratum (implicit + free-form units) is the load-bearing slice for the headline ERA metric; q1 (should-have-changed) is fact-graph metadata supplied to the judge as input rather than a quantity the judge is asked to discover, which explains the near-zero Sonnet--rater $\kappa$ on q1. q2 (did-change) is the dimension that maps to the ERA computation: the judge is asked whether the revision asserts the new fact value, and an ERA credit is granted iff the binary-collapsed verdict is \textsc{correct}. The full grid (including q3a, q3b, and the easy stratum) and the raw annotations are released in \texttt{data/rater/} for third-party re-coding.

\begin{table}[h]
\centering
\small
\begin{tabular}{lllrrl}
\toprule
Pair & Stratum & Dimension & $n$ & $\kappa$ [95\% CI] & one-sided LB \\
\midrule
r1\,$\leftrightarrow$\,r2 & all & q1 & 64 & -0.041 [-0.086, 0.000] & $\geq -0.078$ \\
r1\,$\leftrightarrow$\,r2 & all & q2 & 52 & 0.900 [0.740, 1.000] & $\geq 0.774$ \\
r1\,$\leftrightarrow$\,r2 & hard (implicit+free-form) & q1 & 42 & -0.063 [-0.133, 0.000] & $\geq -0.122$ \\
r1\,$\leftrightarrow$\,r2 & hard (implicit+free-form) & q2 & 30 & 0.933 [0.796, 1.000] & $\geq 0.813$ \\
r1\,$\leftrightarrow$\,sonnet & all & q1 & 64 & 0.077 [-0.045, 0.262] & $\geq -0.038$ \\
r1\,$\leftrightarrow$\,sonnet & all & q2 & 56 & 0.954 [0.841, 1.000] & $\geq 0.865$ \\
r1\,$\leftrightarrow$\,sonnet & hard (implicit+free-form) & q1 & 42 & 0.059 [-0.069, 0.249] & $\geq -0.060$ \\
r1\,$\leftrightarrow$\,sonnet & hard (implicit+free-form) & q2 & 34 & 0.941 [0.815, 1.000] & $\geq 0.833$ \\
r2\,$\leftrightarrow$\,sonnet & all & q1 & 64 & 0.413 [0.109, 0.680] & $\geq 0.165$ \\
r2\,$\leftrightarrow$\,sonnet & all & q2 & 52 & 0.855 [0.669, 1.000] & $\geq 0.709$ \\
r2\,$\leftrightarrow$\,sonnet & hard (implicit+free-form) & q1 & 42 & 0.377 [0.079, 0.648] & $\geq 0.125$ \\
r2\,$\leftrightarrow$\,sonnet & hard (implicit+free-form) & q2 & 30 & 0.871 [0.683, 1.000] & $\geq 0.709$ \\
\bottomrule
\end{tabular}
\caption{Stratified Cohen's $\kappa$ on the 64-item v1.0 human pilot. Bootstrap 95\% CIs use $B=10000$ item-level resamples, seed 4242. The hard stratum (implicit+free-form units) is the load-bearing slice for the headline ERA metric. q1 = should-have-changed; q2 = did-change.}
\label{tab:pilot_kappa_stratified}
\end{table}

\section{Adversarial-baseline gate verdicts}
\label{sec:appendix-gate}

Table~\ref{tab:gate} reports per-rule verdicts of the eight
deterministic adversarial probes at each of the three corpus scales
referenced in Section~\ref{sec:results}.

\begin{table}[h]
\centering
\small
\caption{Adversarial-baseline verdicts at three corpus scales. \texttt{indet.} denotes an intentionally uninformative region, not a passed or failed metric check. The Identity max CDR\_recon threshold was relaxed from $\le 0.05$ to $\le 0.10$ before public release after a calibration sweep absorbed a single Sonnet-judge false-positive on one of $122$ Identity items (mean cdr\_recon across $122$ items = $0.007$); all other thresholds are unchanged.}
\label{tab:gate}
\begin{tabular}{lccc}
\toprule
Rule & Pilot & Scaled & Cascade \\
\midrule
Identity max CDR\_recon ($\le 0.10$)            & \textbf{pass} & \textbf{pass} & \textbf{pass} \\
Identity max CCG ($\le 0.05$)                   & \textbf{pass} & \textbf{pass} & \textbf{pass} \\
Half-Identity max CCG ($\le 0.05$)              & \textbf{pass} & \textbf{pass} & \textbf{pass} \\
Hedge-Bot v2 paired upper CI ($\le +0.05$)      & \textbf{pass} & \textbf{pass} & \textbf{pass} \\
Hedge-Bot v3 paired upper CI ($\le +0.05$)      & \textbf{pass} & \textbf{pass} & \textbf{pass} \\
Truncation-Abuse paired upper CI ($\le +0.05$)  & \textbf{pass} & \textbf{pass} & indet. \\
Three-Quarter-Identity paired upper CI          & \textbf{pass} & indet. & indet. \\
Null max LES ($=0$)                             & \textbf{pass} & \textbf{pass} & \textbf{pass} \\
\bottomrule
\end{tabular}
\end{table}

\section{Concrete examples: opus direct on cascade hard cells}
\label{sec:appendix-examples}

This appendix lists representative model outputs from \texttt{claude-opus-4-7}
under the \emph{direct edit} protocol, the cell with the highest hard-stratum
ERA in v1.0 (0.705, $n{=}461$). Each example shows the original unit, the
model's revised text, and the per-unit verdict. The three categories
illustrate why the headline ERA depends on judge-assisted scoring: the
model's typical correct response is a natural-English paraphrase that falls
outside the canonical alias list, so a regex-only scorer materially
under-reports performance.

\paragraph{Counts.} Across the $461$ hard cells (implicit or free-form
required-update + deletion units on the cascade corpus), units fall into
five judge categories: Stage A satisfied $= 30$ ($6.5\%$), Stage B rescued
$= 295$ ($64.0\%$), genuine failure $= 111$ ($24.1\%$), partial
$= 8$ ($1.7\%$), and deletion-tagged units correctly emitted as absent
$= 17$ ($3.7\%$). Total: $461$. Under the headline ERA rubric, only
Stage A satisfied and Stage B rescued count as unit successes, so the
point estimate reads $(30+295)/461=0.705$; partial, stale, and
deletion-correctly-absent units score $0$. The $17$ deletion-absent units
are semantically successful but uncredited under the v1.0 Stage A + B
rubric -- a known measurement limitation slated for the v1.1 scorer.

\paragraph{(A) Stage A satisfied - model used a canonical alias.}
Uncommon ($6.5\%$ of hard cells). The model happens to substitute back
to a canonical literal (e.g.\ \texttt{small-scale}) or a phrasing
already in \textsc{IMPLICIT\_QUALIFIER\_FORMS}.

\begin{quote}\small\itshape
Edit: \textsc{small-scale} $\to$ \textsc{medium-scale} (95 $\to$ 235).\\
Original: ``This study relies on a dataset, limited to a focused
handful of items, so its conclusions should be understood within that
scope.''\\
Revised: ``This study relies on a \textbf{medium-scale} dataset, so its
conclusions should be understood within that scope.''
\end{quote}

\begin{quote}\small\itshape
Edit: \textsc{medium-scale} $\to$ \textsc{small-scale} (185 $\to$ 85).\\
Original: ``We treat DocAlignmentBench as a benchmark, midway through
the practical size spectrum, complementing larger natural-document
evaluations.''\\
Revised: ``We treat DocAlignmentBench as a \textbf{small-scale}
benchmark, complementing larger natural-document evaluations.''
\end{quote}

\paragraph{(B) Stage B rescued - model used natural paraphrase.}
The dominant category ($64.0\%$). The model substitutes a natural-English
paraphrase that uses the new qualifier's semantic field
(\textit{narrow}, \textit{broader}, \textit{comprehensive}) without
matching the canonical alias list. Stage A returns INCORRECT; the Sonnet
ensemble labels these as \textsc{satisfied} or
\textsc{benign\_paraphrase}.

\begin{quote}\small\itshape
Edit: \textsc{moderate-pool} $\to$ \textsc{comprehensive-pool} (5 $\to$ 8).
Judge: \textsc{benign\_paraphrase}.\\
Original: ``The evaluation, with intermediate coverage of the model
space, provides comparative coverage across systems.''\\
Revised: ``The evaluation, with \textbf{comprehensive} coverage of the
model space, provides comparative coverage across systems.''
\end{quote}

\begin{quote}\small\itshape
Edit: \textsc{moderate-pool} $\to$ \textsc{narrow-pool} (6 $\to$ 4).
Judge: \textsc{benign\_paraphrase}.\\
Original: ``We compare contemporary language models on the benchmark,
with intermediate coverage of the model space.''\\
Revised: ``We compare contemporary language models on the benchmark,
with \textbf{narrow} coverage of the model space.''
\end{quote}

\begin{quote}\small\itshape
Edit: \textsc{moderate-pool} $\to$ \textsc{narrow-pool} (6 $\to$ 4).
Judge: \textsc{benign\_paraphrase}.\\
Original: ``The evaluation, drawing on a workable cross-section of
systems, provides comparative coverage across systems.''\\
Revised: ``The evaluation, drawing on a \textbf{narrow cross-section}
of systems, provides comparative coverage across systems.''
\end{quote}

\paragraph{(C) Genuine failure - model did not propagate.}
$24.1\%$ of hard cells. The model emits the unit unchanged or with a
cosmetic edit that leaves the qualifier semantics intact. The judge
labels these \textsc{stale}.

\begin{quote}\small\itshape
Edit: \textsc{small-scale} $\to$ \textsc{medium-scale} ($90 \to 195$).
Judge: \textsc{stale}.\\
Original: ``Whether findings hold beyond a setup, limited to a focused
handful of items, remains an open question for future benchmarks.''\\
Revised (unchanged): ``Whether findings hold beyond a setup, limited to
a focused handful of items, remains an open question for future
benchmarks.''
\end{quote}

\begin{quote}\small\itshape
Edit: \textsc{narrow-pool} $\to$ \textsc{moderate-pool} (4 $\to$ 7).
Judge: \textsc{stale}.\\
Original: ``The evaluation, with restricted coverage of the model
space, provides comparative coverage across systems.''\\
Revised (unchanged): ``The evaluation, with restricted coverage of the
model space, provides comparative coverage across systems.''
\end{quote}

\paragraph{Implications.}
Two takeaways follow. First, the unit-level distribution of model behavior on hard cells is dominated by natural paraphrase rather than canonical alias matching, so without judge fallback, hard-stratum ERA captures only Stage A's coincidental matches and materially under-reports model capability. Second, the residual genuine failures are mostly cases where the model emits the unit verbatim. This is consistent with cascade dependencies being neither localized enough for literal alias replacement nor cosmetically detectable from the unit alone. The implicit-qualifier strata therefore measure the capability EditPropBench targets: cross-sentence propagation that is invisible to substitution rules but visible to careful semantic checking.

\newpage

\end{document}